\documentclass[runningheads]{llncs}
\usepackage{graphicx}
\usepackage{indentfirst}
\usepackage{cite}
\usepackage{afterpage}
\usepackage{amsmath,amssymb,amsfonts}
\usepackage{algorithmic}
\usepackage{graphicx}
\usepackage{subfigure}
\usepackage{textcomp}
\usepackage[linesnumbered,ruled]{algorithm2e}
\usepackage{array}
\usepackage{float}
\usepackage{color}

\begin{document}
\title{PCGAN-CHAR: Progressively Trained Classifier Generative Adversarial Networks for Classification of Noisy Handwritten Bangla Characters}
%
%
\author{Qun Liu \and
Edward Collier \and
Supratik Mukhopadhyay}
\authorrunning{F. Author et al.}
%
\institute{Louisiana State University, Baton Rouge LA 70803, USA 
\email{qliu14@lsu.edu,ecoll28@lsu.edu,supratik@csc.lsu.edu}
}
\maketitle              
\begin{abstract}
Due to the sparsity of features, noise has proven to be a great inhibitor in the  classification of handwritten characters. To combat this, most techniques perform denoising of the data before classification. In this paper, we consolidate the approach by training an all-in-one model that is able to classify even noisy characters. For classification, we progressively train a classifier generative adversarial network   on the characters from low to high resolution. We show that by learning the features at each resolution independently a trained model is able to accurately classify characters even in the presence of noise. We experimentally demonstrate the effectiveness of our approach by  classifying  noisy versions of MNIST \cite{Karki18},  handwritten Bangla Numeral, and Basic Character datasets \cite{bhattacharya2009handwritten}, \cite{bhattacharya2012offline}.

\keywords{Progressively Training \and General Adversarial Networks \and Classification \and Noisy Characters \and Handwritten Bangla.}
\end{abstract}
\section{Introduction}
Early work in neural networks focused on classification of handwritten characters \cite{lecun1998gradient, lecun1989backpropagation}. Since then,  there has been a lot of research  on  character recognition. While in many cases, text processing  deals directly with the character strings themselves, there are a growing number of use cases for recognizing characters and text in physical real world prints and documents. This  includes processing receipts and bank statements, transcribing books and medical prescriptions,  or translating text. Aside from the large amounts of computer generated text, there are vast quantities of scanned handwritten text that can be processed. Such text is generally collected as images, which invariably introduces some noise (e.g., damaged documents, noise added due to camera motion, etc.). While computer generated text classification might be more stable to noise,  recognition of handwritten text  breaks down with the introduction of noise.

In this work, we build on recent work in adversarial training \cite{nvidia2017progressive} to improve on the state-of-the-art in representing sparse features \cite{yosinski2014transferable, basu2017learning, nvidia2017progressive}. We define sparse representations as noisy, generally compact, representations of signals \cite{huang2006sparse}\cite{boureau2008sparse}. This is the case for many real world images which contain various sources of noise that can distort their true representation. Such noise can easily reduce the quality of classifications and challenge the power of classifiers \cite{noise1}\cite{noise2}. Most algorithms include denoising step for the images before classification \cite{Karki18}, while our approach can directly classify without denoising step due to progressively learn features at increasing resolutions to accurately classify the noisy digits/characters.

\begin{figure*}[t!]
\centering
\includegraphics[width=0.98\textwidth]{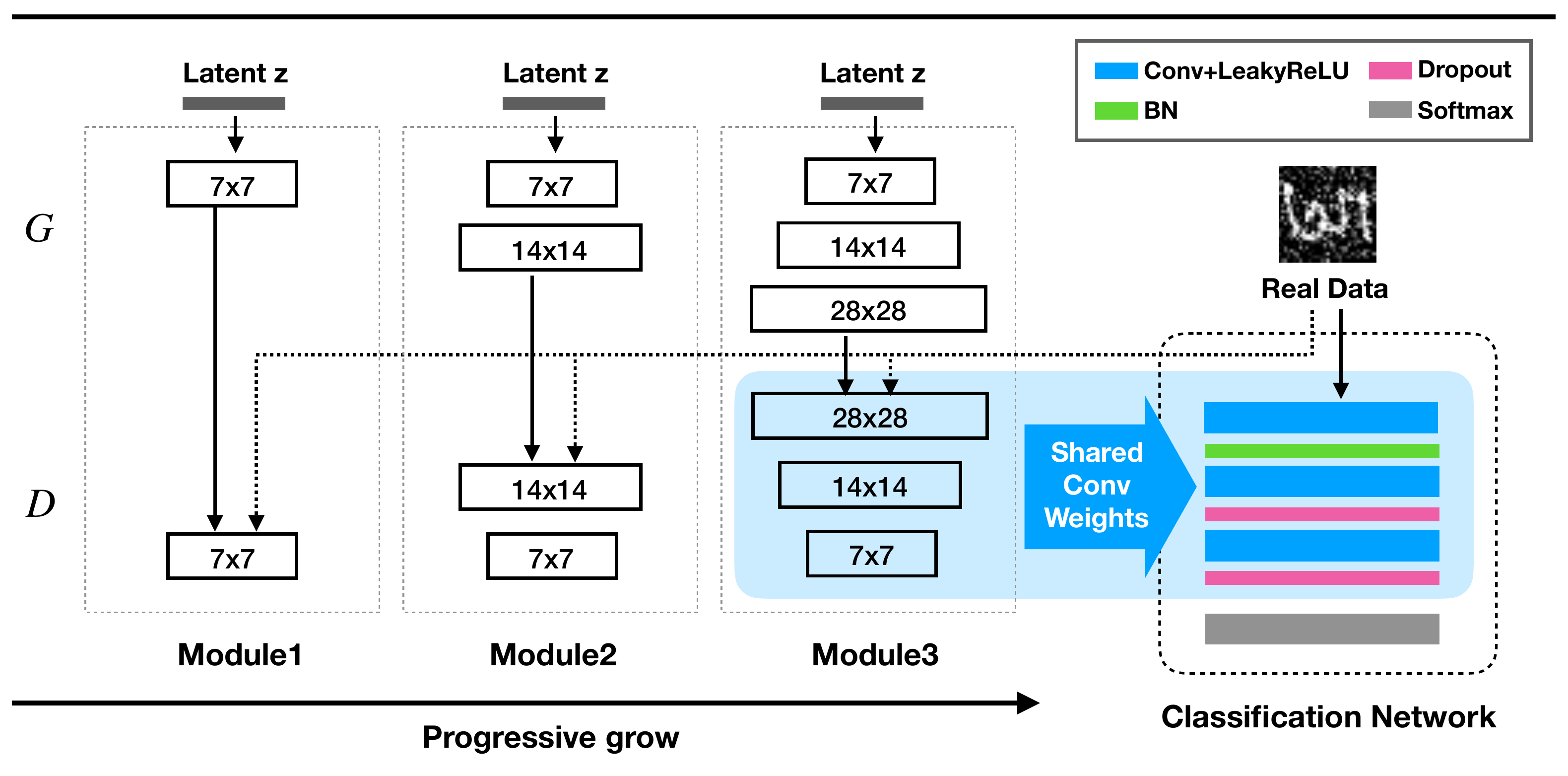}
\caption{Overview of our proposed Progressively Trained Classifier Generative Adversarial Networks (PCGAN-CHAR) architecture.}
\label{fig:bf}
\end{figure*}

Fig.\ref{fig:bf} shows the architecture of our approach. We utilized the progressive technique which is a newly proposed method  \cite{nvidia2017progressive},  for training \emph{Auxiliary Classifier Generative Adversarial Networks (ACGAN)} that has been attractive due to its ability to improve and stabilize the network. It facilitates networks to learn features in a generic to specific manner as the input progresses down the model \cite{yosinski2014transferable}. Low resolution features are more resistant to noise due to their generic nature. By individually learning representations at each resolution, our method is able to leverage  the noise-resistant generic features to make more accurate and better predictions for  noisy handwritten characters to achieve  state-of-the-art performance.

In general, our framework uses a generative adversarial network (GAN) \cite{Goodfellow-GAN} as a basic component for its  noise-resilient ability and   the discriminative power of its discriminator for classification. GANs contain two competing networks, a generator and a discriminator \cite{Goodfellow-GAN}, playing a minmax game. The discriminator \cite{Goodfellow-GAN} tries to discern real samples from fake ones generated by the generator in an attempt to fool the discriminator. Because of this behavior, as one network tries to minimize its own loss,  this in turn maximizes the other network's loss. Generators from GANs have been shown to generate outputs that are almost indiscernible from real samples, while the disrcriminator trained by a GAN has more discriminative power with respect to classification being exposed to both real and fake data, where much of the fake data contains some noise.

We used the discriminator in the Classifier GAN for our Classification Network for the noisy handwritten characters, but to make it  more robust to noise and resolution we novelly adopted the innovative GAN training technique, Progressive growing \cite{nvidia2017progressive}, to our Classifier GAN. In progressive growing each layer of the GAN is trained individually on increasing resolutions. This allows each feature to specialize and simplifies the problem at the individual layers.
   
To the best of our knowledge, this is the first work that designs a Progressive trained Classifier GAN (PCGAN) for retaining the noise-resistant discriminator in a classifier GAN for  robust classification in noisy settings. This paper makes the following contributions.
\begin{itemize}
\item It presents a novel robust noise-resilient classification framework using  progressively trained  classifier general adversarial networks.
\item The proposed classification  framework can directly classify raw noisy data without any preprocessing  steps that include complex techniques such as denoising or reconstruction.
\item It experimentally demonstrates the effectiveness of the framework on the Noisy Bangla Numeral, the Noisy Bangla Characters, and the Noisy MNIST benchmark datasets.
\end{itemize}

       
\section{Related Work}\label{rw}
Handwritten character based datasets have become benchmarks in computer vision research. Handwritten characters contain sparse representations, or features, while also containing significant amounts of noise \cite{basu2017learning}. Early work on classification of handwritten characters focused on dimensionality reduction and denoising. This includes the use of quadtrees \cite{markas1992quad, aref1993decomposing, basu2017learning} and intermediate layers of Convolutional Neural Networks for representations and Deep Belief Networks (DBN) for denoising \cite{basu2017learning}.

Researchers have tried a variety of methods  to solve  the noisy character classification problem. For example, multi-stage approaches have used chain code histogram features to discriminate classes in \cite{bhattacharya2009handwritten}. Similar to this work, increasing resolutions are used to assist in classification. Other multistage approaches have used modified quadratic discriminant function (MQDF) and gradients from neural networks to classify characters from many classes \cite{bhattacharya2012offline}.


\begin{figure*}[t!]
\begin{center}
\subfigure[Samples of Noisy Bangla Characters data with  Added White Gaussian Noise (AWGN) ]{ \includegraphics[width=.28\textwidth]{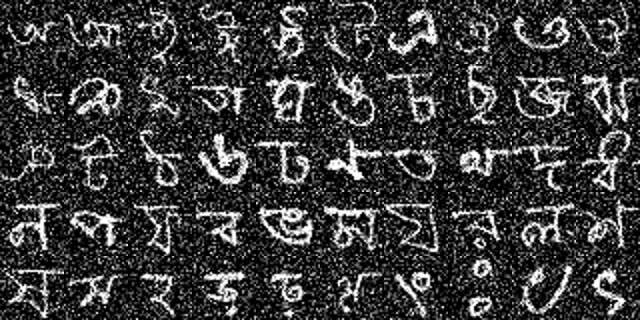}\label{fig:awgn}}
\hspace{5pt}
\subfigure[Samples of Noisy Bangla Characters data with Reduced Contrast with white gaussian noise (Contrast)]{ \includegraphics[width=.28\textwidth]{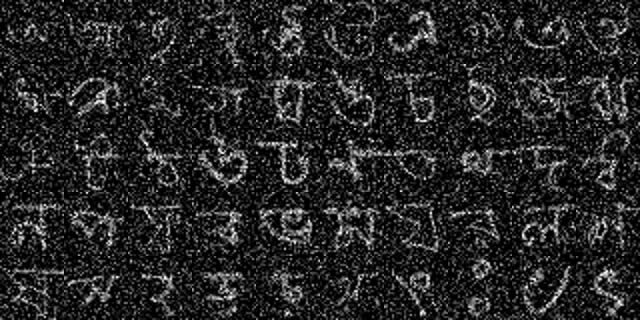}\label{fig:contrast}}
\hspace{5pt}
\subfigure[Samples of Noisy Bangla Characters data with Motion Blurred noise (Motion)]{ \includegraphics[width=.28\textwidth]{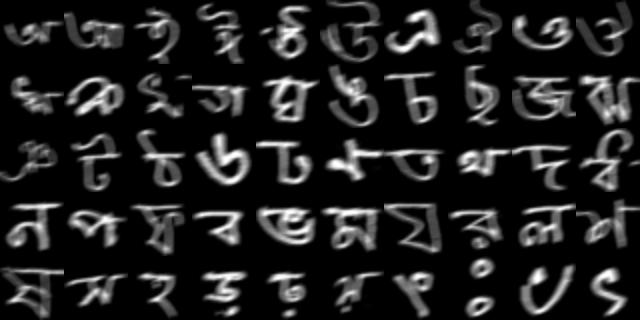}\label{fig:motion}}
\hspace{5pt}
\caption{Sample data for different types of added noise for Noisy Bangla Characters. Three types of noisy data, Added White Gaussian Noise, Reduced Contrast with white gaussian noise, Motion Blurred noise, shown above from top to bottom.}
\label{fig:sample}
\end{center} 
\end{figure*}

Convolutional neural networks (CNNs)  \cite{cactus,deepsat,theoretical,hariharan2015hypercolumns, krizhevsky2012imagenet} have been widely used  in image processing, and are increasingly being used in  character and text classification \cite{Karki18}. The performance of CNNs can be greatly affected by blurry images, where  noise increases the separation between the output and the ground truth in feature space \cite{pathak2016context}. While Euclidean distance has shown to be a decent method for measuring the closeness between two images in feature space, it is difficult to measure the sharpness and quality of images. Through adversarial training, Generative adversarial networks (GANs)  can both imagine structure where there is none, to produce sharp images, and discern real from fake \cite{Goodfellow-GAN}. Progressively growing GANs are an improvement on the GAN architecture that can produce sharp realistic images \cite{nvidia2017progressive}. This innovation is important for GANs operate on high resolution input. When handling high resolution data it can be too easy for the GAN to discriminate between the fake, generally low resolution imagery, and the real,  which are high resolution. Progressively grown GANs learn the resolutions at each layer in isolation by training each layer almost independently, before adding the next layer and training again. This process utilizes transfer learning and the generic to specific learning behaviour exhibited in neural networks \cite{yosinski2014transferable}. Auxiliary Classifier GANs (ACGAN) \cite{google} has introduced class labels in GANs by adding an auxiliary classifier which leveraged model in its prior, their research focused on image synthesis tasks and has shown better performance. The focus of our paper is classification of noisy handwritten (Bangla) characters. 

\section{Methodology}\label{pm}
In this section, we present an overview of our proposed method. We separate our methodology into three subsections; Generative Adversarial Networks, Progressively Trained  Classifier GAN, and Classification Network. In the Generative Adversarial Network section, we outline the formulation of our GAN objective function. We describe the progressive growing structure of the GAN in Section \ref{prog}, along with  a general training algorithm. Details about the Classification Network are  provided in Section \ref{class}.

\subsection{Generative Adversarial Network}

A Generative Adversarial Network (GAN) \cite{Goodfellow-GAN} comprises of two networks:  a generator and a discriminator. The generator $G$  takes  as input a random noise vector $z$   and outputs a fake image $G(z)$. It learns a mapping function, $z \rightarrow y$, between the latent space $z$ and the feature space defined by the task $y$. The discriminator $D$ takes as input  $x$ either  a real image or a fake image produced  by $G$  to calculate class probabilities. It attempts to learn a mapping between feature space $y$ and a discriminatory space $q$, $y \rightarrow q$. The goal is to map all true $y$ to the positive class while all fake $y_f$ to the negative class.

To learn these mappings the generator $G$ plays with the discriminator $D$   a two-player min/max game, $\mathit{min}_G \:\mathit{max}_D\: L(G, D)$, with a loss $L(G, D)$.  The objective function \cite{Goodfellow-GAN} is defined  as follows. 
\begin{equation}
\begin{aligned}
\mathit{min}_G \:\mathit{max}_D\:\:L(G, D) = &\mathbb{E}_{y}[logD(y)] \\
& + \mathbb{E}_{x,z}[log(1 - D(x|G(x|z)))]
\end{aligned}
\end{equation}
where the discriminator $D$ will try to maximize the log-likelihood which is the first term, while the generator $G$ try to minimize the second term.

The classifier GAN \cite{google}  takes a class label $l$ and a random noise vector $z$ to generate   fake images $X_f$ through its generator $G$, denoted as $G(l, z)$. The real images $X_r$  are the training images in the noisy dataset under consideration.  The objective function of the classification GAN comprises of two components each involving two log-likelihood $L_1$ and $L_2$.  The log-likelihood $L_1$  involves   the  conditional probability distribution $P(\mathit{guess} \mid x)$ where  the input $x$ can be a fake image  $X_f$ (generated by the generator) or a real image  $X_r$ (a training image from the noisy dataset under consideration) and $\mathit{guess}$ can take two values $\mathit{fake}$ or $\mathit{real}$.  Formally \cite{google}, 
\begin{equation}
\begin{aligned}
L_1 = \mathbb{E}[\log P(\mathit{real} \mid X_r)] + \mathbb{E}[\log P(\mathit{fake} \mid X_f)].
\end{aligned}
\end{equation}
The log-likelihood $L_2$ involves the conditional probability distribution $P(l \mid x)$  where $l$ is the class label  of $x$ and $x$ is a real image $X_r$ or a fake image $X_f$. Formally, 
\begin{equation}
\begin{aligned}
L_2 = \mathbb{E}[\log P(l \mid X_r)] + \mathbb{E}[\log P(l \mid X_f)].
\end{aligned}
\end{equation}
The generator $G$ will  try to maximize $L_2 - L_1$, but the discriminator $D$ will try to maximize $L_1 + L_2$.

\subsection{Progressively Trained Classifier GAN}\label{prog}

Progressive growing is a recent development from \cite{nvidia2017progressive} that uses transfer learning to improve the quality of the learned models. Training is performed individually for the layers of the generator $G$ and the discriminator $D$. New mirroring layers are added to $G$ and $D$ before a new training iteration is run. This increases the spatial resolution of the output image for each layer progressively added. Layers in $G$ and $D$ become more specialized to  spatial resolution, resulting in them learning finer features. In addition to making the layers more specialized, progressively growing also simplifies the problem at each layer. This creates a more stable generative model with finer outputs. 

Deep neural networks learn features in a low resolution to high resolution manner, or generic to specific. Progressive growing takes advantage of this behavior by transferring the weights learned from all previous training iterations to identical layers for the next step. Each training iteration then only contains one untrained layer, the newest layer, allowing for features at each resolution to be learned independently and in relative isolation from the other layers. The discriminator grows in parallel with the generator with each layer learning to discriminate specific resolutions.

Most progressive GANs are designed with generative tasks in mind, putting a focus on the generator. In our work, however, we use the well trained discriminator that is the end result of progressive training. Each layer of the discriminator specializes at specific resolutions allowing it to learn more fine-grain features. Additionally, the discriminator has seen a wide range of samples produced by the generator as it learns; from noisy to sharp. We call this discriminator ``well trained" because it is more robust to noise and sparse features, characteristics inherent in handwritten characters.

We show the general  training algorithm for our proposed framework for progressively  training the classifier GAN   in Algorithm \ref{alg1} for a given number of progressive stages. As seen in the Algorithm, we first initialize the progressive stages (indicated as  modules in Fig.\ref{fig:bf}), for learning essential features at different resolutions. The input resolution for  the discriminator increases from $7 \times 7$ for the first module   to $28 \times 28$ for the third module as seen in  Fig.\ref{fig:bf}. After initializing a  module, our algorithm begins training the GAN using the  normal training procedure. Since our GAN is used for classification purposes, unlike other variants of GANs used to synthesize high quality images, which mainly focus on  the generator,  we  focus on the  discriminator and aim to improve its classification ability. Similar to  ACGAN \cite{google},  we add an auxiliary classifier  to  the discriminator to compute class labels, apart from  using a  binary classifier   for discerning if an image is fake or real. The latter classifier corresponds to the loss $\mathcal{L}_{discern}$ as shown in Algorithm \ref{alg1}; this loss is used to enhance the robustness of the discriminator by learning better representations of variations within a class \cite{hariharan2017low}. After a module is trained through epochs, the weights in its generator and discriminator are transferred to the next module (as seen in Fig.\ref{fig:bf}, the weights from $\mathit{Module}_i$ is transferred to  $\mathit{Module}_{i+1}$, with $i \geq 1$, with the number of layers increasing as we progress from the $i$th module to the $i+1$th module for augmenting resolution). This method of progressive training continues until training for the last module has converged.

\subsection{Classification Network}\label{class}
After being progressively trained, the discriminator of the classifier GAN  has learned   the input space in such a way that the lower layers specialize on low resolutions while the higher layers specialize on high resolutions.   Progressive training results in a stabilized  discriminator that has  learned the essential features from noisy data at multiple resolutions, resulting in better classification performance. As we can see in Fig.\ref{fig:bf},  as we go from the $i$th module to the $i+1$th module, the number of layers increases. The input resolution  for discriminator increases from $7 \times 7$ in the first module to $28 \times 28$ in the third module.  The  weights of trained discriminator  in Module 3 are transferred to the  classification network, which is a convolutional neural network as described below.    To further improve the performance of the classification network, we finetuned the softmax layer. 
The details of the  classification network are shown  in Fig.\ref{fig:bf}. It includes three convolutional layers,  each  with LeakyReLU activation function.   Two dropout layers and a batch normalization layer are added to the classification network as showed in Fig.\ref{fig:bf}. Finally, a softmax layer added as the last layer for classification.

\begin{algorithm}[t]
\caption{General Training Algorithm} \label{alg1}
\For{number of $modules$}
{
    Initialize($module$) \par
    \For{epoch=1,2,...,K}
    {
        \For{number of batches}
        {
            $B_r$ $\gets$ Sample a batch of $n$ real images with labels \par
            $B_z$ $\gets$ Sample a batch of $n$ random vectors \par
            $B_l$ $\gets$ Sample a batch of $n$ random labels \par
            Update the parameters in the discriminator regarding gradients,
            \begin{align*}
            \nabla_{\theta_d} \dfrac{1}{2n} \mathlarger{\sum}_{r \in B_r, z \in G(B_z, B_l)} \mathcal{L}_{discern}(r, z) \\
            + \dfrac{1}{n} \mathlarger{\sum}_{r \in B_r} \mathcal{L}_{class}(r)
            \end{align*}
            
            $B_z$ $\gets$ Sample a batch of $2n$ random vectors \par
            $B_l$ $\gets$ Sample a batch of $2n$ random labels \par
            Set discriminator trainable to false, update the parameters in the generator regarding gradients,
            \begin{align*} 
            \nabla_{\theta_g} \dfrac{1}{2n} \mathlarger{\sum}_{z, z' \in G(B_z, B_l)} \mathcal{L}_{discern}(z, z') \\
            + \dfrac{1}{2n} \mathlarger{\sum}_{z, z' \in G(B_z, B_l)} \mathcal{L}_{class}(z, z')
            \end{align*} 
            }
    }
    Transfer weights to Next($module$)
}
\end{algorithm}

\section{Experimental Evaluation}\label{exp}
The experiments mainly focus on Indian handwritten digits and characters datasets \cite{Karki18}, namely, Noisy Bangla Numeral and Noisy Bangla Characters, are publicly available datasets we downloaded from online\footnote{https://en.wikipedia.org/wiki/List\_of\_datasets\_for\_machine\-learning\_research\#Handwriting\_and\_character\_recognition}, it provided a training dataset and a test dataset. For a comparative study, we also conducted experiments on a noisy version of a commonly used handwritten digits dataset, Noisy MNIST Dataset \cite{Karki18}. The original non-noisy versions of the Bangla Numeral and Character datasets are from \cite{bhattacharya2009handwritten, bhattacharya2012offline}.  We consider three different versions of each dataset, the first with Added White Gaussian Noise (AWGN),  the second with Reduced Contrast with white Gaussian noise (Contrast), and  the third with Motion Blurred noise (Motion). Sample data from each of the three different versions of the noisy Bangla character dataset are shown in  Fig.\ref{fig:sample}. 

\subsection{Datasets}

\textbf{Noisy Bangla Numeral} has three different versions, each with different type of noise added, AWGN, Contrast, and Motion. For each version, there are 10 classes of Bangla Numerals with a total of 23330 black  and white images with image size $32\times32$.

\textbf{Noisy Bangla Characters} contains 76000 black  and white images for $50$ classes of Bangla Characters with image size $32\times32$, in each version. There are  three different versions one for each type of added noise as stated above.

\textbf{Noisy MNIST Dataset} is the  same as the  original MNIST dataset except  for added noise. Again, there are three different versions, one for each of the three types of noise considered.  Each version  contains 10 classes with a total of 70000 black and  white images with image size $28\times28$.

\subsection{Implementation details}

We used an  auxiliary classifier GAN (ACGAN) \cite{google} as the classifier GAN in our framework.  We considered $28\times28$ (the resolution of the noisy MNIST dataset)  as the input resolution for our framework.  The images in the noisy Bangla  dataset were resized to $28 \times 28$ before being input to our framework. 

The architecture of generators is the reverse of that of the discriminators (see Fig.\ref{fig:bf}).  
Specifically, for  the generators, we used a dense layer to transform an input (latent $z$) to  a  format that corresponds to the input of the  convolutional transpose layers for generating multiresolution images. For $\mathit{Module}_1$,  after the first convolutional transpose layer, we used a batch normalization layer and another convolutional transpose layer to transform  the feature maps to an output image with resolution  $7\times7$ using a filter with kernel size $1\times1$. A similar configuration is used  generating output images with resolutions  $14\times14$ and $28\times28$  respectively for $\mathit{Module}_2$ and $\mathit{Module}_3$ during the  progressive training procedure. For the discriminator, during progressive training, we started from the input resolution of $7\times7$ ($\mathit{Module}_1$).  The real images (i.e., the training images from the noisy dataset under consideration) are downscaled to $7 \times 7$.      The downscaled  real images are combined   with fake images  produced by the  generator and are fed to the discriminator for feature extraction and classification. The discriminator not only computes the class label but also discerns fake images from real ones. Similar downscaling and combination operations are performed before feeding  inputs to the discriminators for the second and the third modules. 

 During the progressive training,  the GAN learns, in addition to other features,  low resolution features that are more tolerant to noise, being generic in nature. While the GAN learns low resolution as well as noisy features, it never learns to denoise an image. At no point,  does our framework produce  a denoised image or learns the representation of a denoised image.  Instead, low-resolution features that are not disrupted by noise are learned though progressive training.  By individually learning representations at each resolution, our method is able to leverage the noise-resistant generic  low resolution features to provide better  classification performance even for noisy character data. This produces a robust discriminator that can classify noisy  handwritten characters.  Thus, in our framework, there is no preprocessing step that denoises the input explicitly or implicitly. 

\begin{figure*}[p]
\centering
\subfigure[Classification accuracy on three types of Noisy Bangla Numeral.]{
    \includegraphics[width=0.6\textwidth, keepaspectratio]{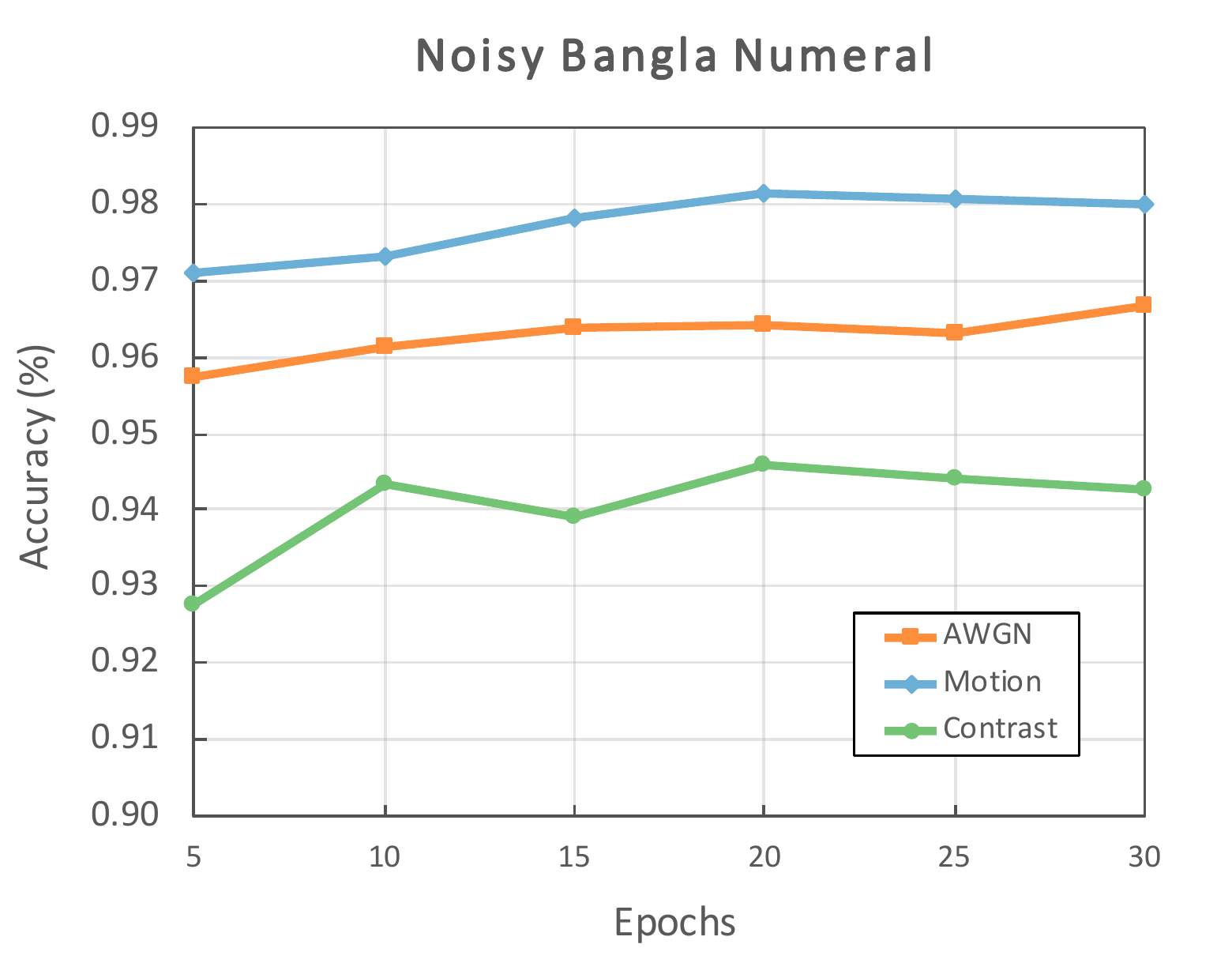}
    \label{fig:numeral}}
\subfigure[Classification accuracy on three types of Noisy Bangla Characters.]{
    \includegraphics[width=0.6\textwidth, keepaspectratio]{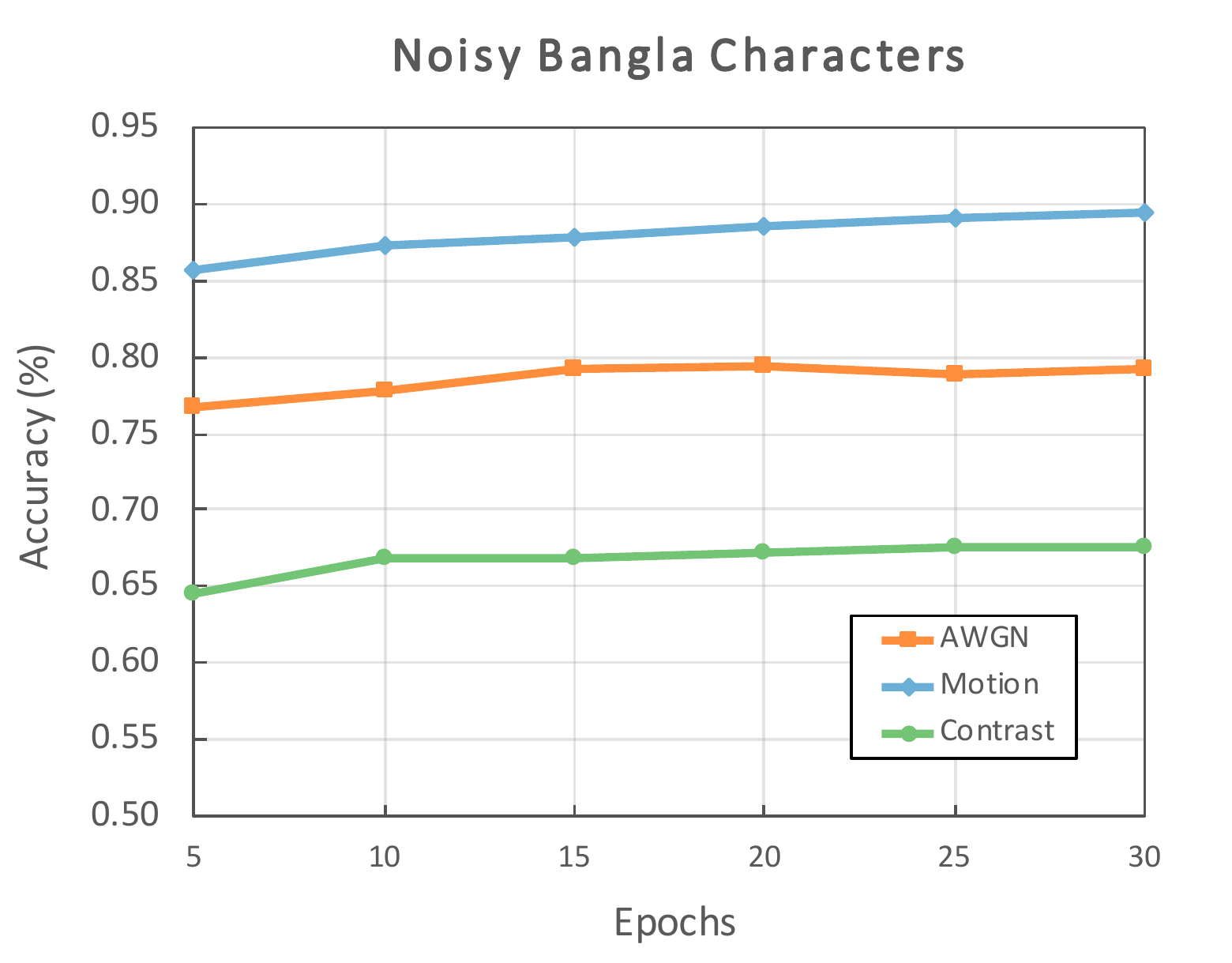}       
    \label{fig:characters}}
\subfigure[Classification accuracy on three types of Noisy MNIST Dataset.]{
    \includegraphics[width=0.6\textwidth, keepaspectratio]{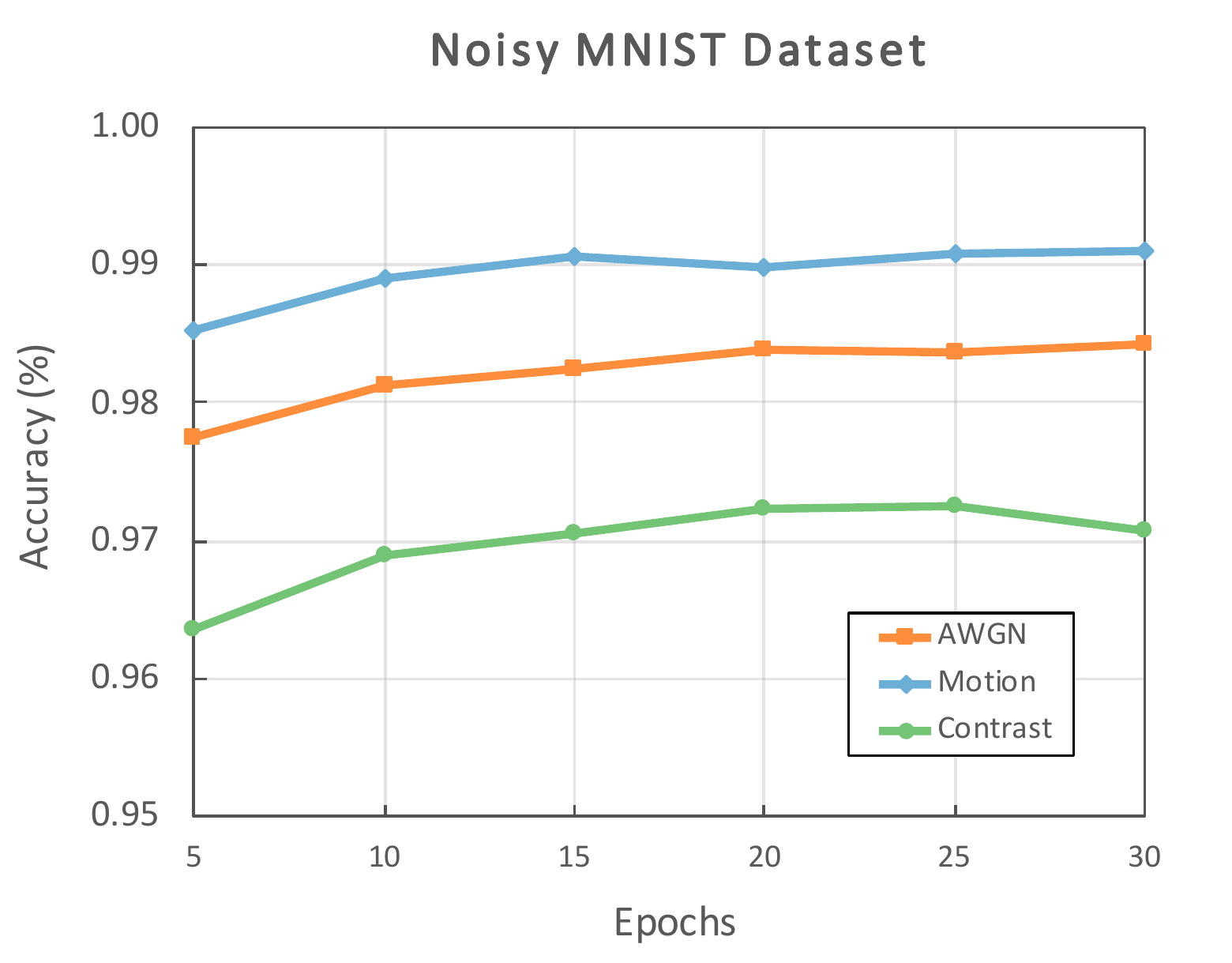} \label{fig:mnist}}
\caption[Optional caption for list of figures]{The classification accuracy of our approach. Our approach has been evaluated on datasets of Noisy Bangla Numeral, Noisy Bangla Characters, and Noisy MNIST with three types of added noise, AWGN, Motion, Contrast.}
\label{fig:train_results}
\end{figure*}

\subsection{Evaluation}
We have evaluated our  framework on Noisy Bangla Numeral, Noisy Bangla Characters, and Noisy MNIST datasets with respect to classification accuracy as shown in Tables \ref{tb_accu_numeral}, \ref{tb_accu_characters}, and \ref{tb_accu_mnist}, respectively. Fig.\ref{fig:train_results} shows how the accuracy  varies with the number of training epochs for Noisy Bangla Numeral (Fig.\ref{fig:numeral}), Noisy Bangla Characters (Fig.\ref{fig:characters}, and Noisy MNIST (Fig.\ref{fig:mnist}) datasets. 

For each version (corresponding to each of the three types of added noise) of each dataset, we used the provided training  and testing dataset  splits \cite{Karki18}.   As we can see in  Table \ref{tb_accu_numeral}, for  the Noisy Bangla Numeral dataset, our approach achieved the best performance (in terms of accuracy): \textbf{96.68\%} on AWGN noise, \textbf{98.18\%} on Motion noise, and \textbf{94.60\%} on Contrast noise, which surpassed the second best ones by \textbf{1.22\%}, \textbf{1.13\%}, \textbf{1.75\%}, respectively. In  Fig.\ref{fig:numeral}, one can see  that the classification accuracy of our framework remains almost the same on AWGN noise with increasing number of epochs. For the same dataset, but with added  Motion noise, the classification accuracy initially increases before stabilizing.  In case of added Contrast noise, for the same dataset, the classification accuracy remains relatively unstable  with increasing number of epochs.


In Table \ref{tb_accu_characters}, for Noisy Bangla Characters, our framework obtained better performance than the state-of-the-art in the case of added AWGN noise (\textbf{79.85\%}) and added Motion noise (\textbf{89.54\%})  surpassing the second best by \textbf{3.11\%} and \textbf{5.95\%}, respectively. In the case of added Contrast noise, our framework achieved an accuracy of  68.41\%, which is (\textbf{-1.25\%}) slightly less than the state-of-the-art  69.66\%. 

The performances of all methods on the Noisy Bangla Characters are much worse than  those on  the Noisy Bangla Numeral and  the Noisy MNIST Datasets.  Noisy Bangla Characters are relatively harder to classify than the other two datasets as this dataset  has 50 classes versus 10 classes  for each  of the  other two  datasets. Additionally, among  the three types of added noises, all methods have relatively poor performances on Contrast noise compared to their performances on AWGN noise and Motion noise. In  Fig.\ref{fig:characters}, we can see that the classification accuracy of our framework remains relatively stable with increasing number of epochs on all the three  types of added noise. 


\begin {table}[b!]
\caption {Comparison of classification accuracy (\%) on three types of Noisy Bangla Numeral}
\begin{center}
\begin{tabular}{ p{4.5cm}|c|c|c }
 \hline
 Methods & AWGN & Motion & Contrast\\
 \hline
 Basu et al. \cite{basu2017learning} & 91.34 & 92.66 & 87.31 \\
 Dropconnect \cite{Karki18}  & 91.18 & 97.05 & 85.79 \\
 Karki et al. (w/o Saliency) \cite{Karki18}  & 95.08 & 94.88 & 92.60 \\
 Karki et al. (Saliency) \cite{Karki18}  & 95.46 & 95.04 & 92.85 \\
 \textbf{PCGAN-CHAR (Ours)} & \textbf{96.68} & \textbf{98.18} & \textbf{94.60} \\
 \hline
\end{tabular}
\label{tb_accu_numeral}
\end{center}
\end{table}

\begin {table}[b!]
\caption {Comparison of classification accuracy (\%) on three types of Noisy Bangla characters}
\begin{center}
\begin{tabular}{ p{4.5cm}|c|c|c }
 \hline
 Methods & AWGN & Motion & Contrast\\
 \hline
 Basu et al. \cite{basu2017learning} & 57.31 & 58.80 & 46.63 \\
 Dropconnect \cite{Karki18}  & 61.14 & 83.59 & 48.07 \\
 Karki et al. (w/o Saliency) \cite{Karki18}  & 70.64 & 74.36 & 58.89 \\
 Karki et al. (Saliency) \cite{Karki18}  & 76.74 & 77.22 & \textbf{69.66} \\
 \textbf{PCGAN-CHAR (Ours)} & \textbf{79.85} & \textbf{89.54} & 68.41 \\
 \hline
\end{tabular}
\label{tb_accu_characters}
\end{center}
\end{table}

\begin {table}[b!]
\caption {Comparison of classification accuracy (\%) on three types of Noisy MNIST dataset}
\begin{center}
\begin{tabular}{ p{4.5cm}|c|c|c}
 \hline
 Methods & AWGN & Motion & Contrast\\
 \hline
 Basu et al. \cite{basu2017learning} & 90.07 & 97.40 & 92.16 \\
 Dropconnect \cite{Karki18}  & 96.02 & 98.58 & 93.24 \\
 Karki et al. (Saliency) \cite{Karki18}  & 97.62 & 97.20 & 95.04 \\
 \textbf{PCGAN-CHAR (Ours)} & \textbf{98.43} & \textbf{99.20} & \textbf{97.25} \\
 \hline
\end{tabular}
\label{tb_accu_mnist}
\end{center}
\end{table}

We also conducted experiments on the Noisy MNIST Dataset.  The classification accuracies  are shown in Table \ref{tb_accu_mnist}. In  Table \ref{tb_accu_mnist}, it can be seen that the accuracy obtained by our framework exceeds the state-of-the-art by \textbf{0.81\%} in the case of added AWGN noise, by \textbf{0.62\%} in the case of added  Motion noise, and by \textbf{2.21\%} in the case of added  Contrast noise, yielding best classification accuracies of \textbf{98.43\%}, \textbf{99.20\%}, and \textbf{97.25\%}, respectively. In Fig.\ref{fig:mnist} for  the Noisy MNIST Dataset, we can see the classification accuracies of our framework already  surpassed the state-of-the-art  after 5 epochs for all the three types of added noise. 

To understand the statistical significance of the performance improvements obtained by our framework over \cite{Karki18}, we used McNemar’s test   (since our framework and \cite{Karki18} had same test datasets).  Following are the results of the McNemar’s tests. 

For noisy Bangla numeral with AWGN noise added: $\chi^2  = 47.02$, $df = 1$, $p<7.025e-12$; with reduced contrast and white Gaussian noise: $\chi^2 = 68.014$, $df = 1$, $p<2.2e-16$; here $df$ represents degrees of freedom.

For noisy Bangla characters with added AWGN: $\chi^2  = 398$, $df = 1$, $p<2.2e-16$.

 For noisy MNIST with added AWGN: $\chi^2  = 79.012$, $df = 1$, $p<2.2e-16$; with reduced contrast and white Gaussian noise: $\chi^2  = 219$, $df = 1$, $p<2.2e-16$.

Based on the results of the McNemar tests, the improvements obtained over \cite{Karki18}, even in the case of AWGN and contrast variations are statistically significant. 


 The discriminator in our framework is trained  to learn representations  progressively from lower resolution to higher. Each layer of the discriminator
specializes at specific resolutions allowing it to learn more
fine-grain features. Lower resolution features are more resistant to noise due to their generic nature. 
Since the discriminator in our framework  has been trained with a combination of fake images produced by the generator and real images belonging to the  noisy dataset under consideration, it learned better representations of the variability within the classes.   This explains the robustness of our framework to noise and sparse features. 

\section{Conclusion}\label{con}

In this paper, we presented a novel robust noise-resilient classification framework for noisy handwritten (Bangla) characters  using  progressively trained  classification general adversarial networks.
 The proposed classification  framework can directly classify raw noisy data without any preprocessing.
 We  experimentally demonstrated the effectiveness of the framework on the Noisy Bangla Numeral, the Noisy Bangla  Basic Characters, and the Noisy MNIST benchmark datasets.
%
%
%
\bibliographystyle{splncs04}
\bibliography{main}

\end{document}